\documentclass[letterpaper]{article} % DO NOT CHANGE THIS
\usepackage{aaai24}  % DO NOT CHANGE THIS
\usepackage{times}  % DO NOT CHANGE THIS
\usepackage{helvet}  % DO NOT CHANGE THIS
\usepackage{courier}  % DO NOT CHANGE THIS
\usepackage[hyphens]{url}  % DO NOT CHANGE THIS
\usepackage{graphicx} % DO NOT CHANGE THIS
\usepackage{natbib}  % DO NOT CHANGE THIS AND DO NOT ADD ANY OPTIONS TO IT
\usepackage{caption} % DO NOT CHANGE THIS AND DO NOT ADD ANY OPTIONS TO IT
\usepackage{algorithmic}
\usepackage{threeparttable}
\usepackage{diagbox}
\usepackage{multirow}
\usepackage{latexsym, amsmath, amssymb, amsfonts, mathrsfs}
\usepackage{xspace}
\usepackage[T1]{fontenc}
\usepackage{float}
\usepackage{color}
\usepackage{url}
\usepackage[ruled,linesnumbered,noresetcount,vlined]{algorithm2e}
\usepackage{todonotes}
\usepackage{picinpar}
\usepackage{multirow}
\usepackage{subcaption}
\usepackage{amsthm}
\usepackage{booktabs} 
\usepackage{diagbox}
\allowdisplaybreaks
\usepackage{pgfplots}

\newcommand{\fml}[1]{\mathcal{#1}}

\newcommand{\beitemize}{\begin{list}{$\bullet$}{\topsep=1.5pt \parsep=0pt \itemsep=1pt \leftmargin=1em }} 
\newcommand{\enitemize}{\end{list}}

\newcommand{\beenumerate}{\hspace{-0.5in} \begin{enumerate}\topsep=1pt \parsep=0pt \itemsep=-3pt} \newcommand{\enenumerate}{\end{enumerate}}

\newcommand{\belist}{\begin{list}{$\bullet$}{\topsep=1.5pt \parsep=0.5pt \itemsep=1pt \leftmargin=2.25em \labelwidth=1.0em \labelsep=0.5em \partopsep=1.5pt}} 
\newcommand{\enlist}{\end{list}}

\newtheorem{definition}{{\bf Definition}}

\newtheorem{example}{{\bf Example}}

\newboolean{includeMemo}
\setboolean{includeMemo}{true} 

\newcommand{\memoside}[1]{\ifthenelse{\boolean{includeMemo}}{\todo[caption={},color=green!20!]{{\footnotesize #1}}}}
\newcommand{\memo}[1]{\ifthenelse{\boolean{includeMemo}}{\todo[inline,caption={},color=green!20!]{#1}}}
\newcommand{\memob}[1]{\ifthenelse{\boolean{includeMemo}}{\todo[inline,caption={},color=blue!20!]{#1}}}

\newcommand{\xhdr}[1]{\vspace{5pt}\noindent\textbf{#1 }}
\newcommand{\ignore}[1]{}

\newcommand{\squishlist}{
\begin{list}{{{\small{$\bullet$}}}}
{\setlength{\itemsep}{3pt}      
\setlength{\parsep}{3pt}
\setlength{\topsep}{3pt}       
\setlength{\partopsep}{3pt}
\setlength{\leftmargin}{1em} 
\setlength{\labelwidth}{1em}
\setlength{\labelsep}{0.5em} } }
\newcommand{\squishend}{  \end{list}}

\newcommand{\squishenum}{
\begin{list}{$\bullet$}{ 
    \setlength{\itemsep}{1pt}
    \setlength{\parsep}{0pt}
    \setlength{\topsep}{1.5pt}
    \setlength{\partopsep}{0pt}
    \setlength{\leftmargin}{2em}
    \setlength{\labelwidth}{1.5em}
    \setlength{\labelsep}{0.5em} } }

\usepackage{newfloat}
\usepackage{listings}
\DeclareCaptionStyle{ruled}{labelfont=normalfont,labelsep=colon,strut=off} % DO NOT CHANGE THIS
\floatstyle{ruled}
\newfloat{listing}{tb}{lst}{}
\floatname{listing}{Listing}
\title{Approximating Human Models During Argumentation-based Dialogues}
\author{
    Yinxu Tang,
    Stylianos Loukas Vasileiou,
    William Yeoh
}
\affiliations{
    Washington University in St. Louis\\
    \{t.yinxu,v.stylianos, wyeoh\}@wustl.edu
}

\usepackage{pgfplots}
\pgfplotsset{compat=1.18}

\begin{document}
\maketitle

\begin{abstract}
Explainable AI Planning (XAIP) aims to develop AI agents that can effectively explain their decisions and actions to human users, fostering trust and facilitating human-AI collaboration. A key challenge in XAIP is model reconciliation, which seeks to align the mental models of AI agents and humans. While existing approaches often assume a known and deterministic human model, this simplification may not capture the complexities and uncertainties of real-world interactions. In this paper, we propose a novel framework that enables AI agents to learn and update a probabilistic human model through argumentation-based dialogues. Our approach incorporates trust-based and certainty-based update mechanisms, allowing the agent to refine its understanding of the human's mental state based on the human's expressed trust in the agent's arguments and certainty in their own arguments. We employ a probability weighting function inspired by prospect theory to capture the relationship between trust and perceived probability, and use a Bayesian approach to update the agent's probability distribution over possible human models. We conduct a human-subject study to empirically evaluate the effectiveness of our approach in an argumentation scenario, demonstrating its ability to capture the dynamics of human belief formation and adaptation. 
\end{abstract}

\section{Introduction}

The increasing integration of AI systems into real-world applications has underscored the critical need for transparency and trust in human-AI interactions. In this landscape, Explainable AI Planning (XAIP) has emerged as a pivotal area of focus \citep{Fox2017ExplainableP}, propelled by its promise to develop AI agents capable of explaining their decisions and actions in a manner comprehensible to human users. Central to XAIP is the concept of \textit{model reconciliation} \citep{chakraborti2017plan}, a process aimed at aligning the mental models of AI agents and human users to facilitate better understanding and communication. These mental models are typically encoded using planning paradigms \citep{sreedharan2021using,sreedharan2021foundations} or logical formalisms \citep{son2021model,vas21,vasileioulogic,vasileioua2023please}.

However, a common assumption in most XAIP-related work has been that the AI agent possesses a known and deterministic model of the human user, which is used in the agent's deliberative processes. This simplistic approach may fail to capture the intricate complexities of real-world interactions, as humans often hold beliefs with varying degrees of certainty, and their beliefs evolve dynamically over time. Such simplifications can lead to significant misalignments between AI agents and human users, as the agents might base their decisions or explanations on an inaccurate or incomplete understanding of the human's mental model. Consequently, this can result in decreased human trust and engagement with AI agents, undermining the fundamental goals of the human-AI interaction community and hindering the development of human-compatible AI systems \citep{russell2019humancompatible}.

To address this challenge, we propose a novel approach that enables AI agents to adapt their decisions and explanations based on a more nuanced and dynamic understanding of human mental states. We relax the assumption of a deterministic human model and instead posit that the AI agent maintains a probabilistic representation of the human's knowledge. This probabilistic human model is learned and updated dynamically through ongoing interactions, allowing the agent to refine its understanding of the human's mental state over time. To facilitate this learning process, we introduce a framework that learns and updates the probabilistic human model through argumentation-based dialogues \citep{gordon1994,prakken2006formal,parsons2003properties,Rago_23,vasileiou2023dr}. Argumentation provides a natural and expressive mechanism for the agent and the human to exchange information, beliefs, and justifications, allowing for a rich and dynamic interaction. Our learning framework incorporates two complementary update mechanisms: a \textit{trust-based update} and a \textit{certainty-based update}.

The trust-based update mechanism allows the AI agent to adjust its probabilistic human model based on the human's expressed trust in the agent's arguments. Specifically, when the agent presents an argument, the human may not fully accept it as true, but rather evaluate it based on their trust in the agent's argument. We capture this notion of trust-based uncertainty using a trust value $\tau(A_i) \in [0,1]$ associated with each agent argument $A_i$, where higher values indicate greater trust. To capture the relationship between the human's trust in the agent's argument and the probability of that argument, we employ a probability weighting function inspired by prospect theory \citep{tversky1992advances}. This function maps the human's trust $\tau(A_i)$ in the agent's argument to a probability $p(A_i)$ in a way that accounts for the psychological biases humans exhibit when assessing probabilities under uncertainty.

The certainty-based update mechanism focuses on the human's expressed certainty in their own arguments. When the human puts forward an argument, they may express some degree of uncertainty about it, represented by a probability $p(A_j) \in [0,1]$. This probability reflects the human's confidence in their own argument $A_j$, based on factors such as their background knowledge, reasoning process, and awareness of potential counterarguments. 

To update the (probabilistic) human model, we use a Bayesian update mechanism. Particularly, we update the agent's probability distribution over possible human models based on the uncertainties associated with the arguments exchanged during the dialogue. At each timestep, when an argument is presented by either the agent or the human, we perform a general update on the probability distribution that increases the probability of the models consistent with the argument, weighted by the argument's associated probability, and decreases the probability of the models inconsistent with the argument.

Finally, to assess the framework's ability to approximate human models through argumentation-based dialogues, we conduct a human-user study that simulates a decision-making scenario. Our findings demonstrate the feasbility of our approach to dynamically approximate a human model as a probability distribution, leading to increased trust and satisfaction among participants.
% and an increase in human trust as the dialogue progresses.

The main contributions of this paper are as follows:
\squishlist
    \item We propose a novel framework for learning and updating a probabilistic human model through argumentation-based dialogues, where we incorporate trust-based and certainty-based update mechanisms.

    \item We conduct a human-user study to empirically evaluate the effectiveness of our approach in decision-making scenario, demonstrating its ability to capture the dynamics of human belief formation and adaptation.
\squishend

% The rest of this paper is structured as follows. In Section \ref{sec:related_work}, we review related work on XAIP, model reconciliation, and argumentation-based dialogues. Section \ref{sec:framework} presents our formal framework for learning and updating a probabilistic human model through argumentation-based dialogues. In Section \ref{sec:experiment}, we describe our human-subject study and discuss the results and implications of our empirical evaluation. Finally, Section \ref{sec:conclusion} concludes the paper and outlines directions for future research.

\section{Related Work}

We situate our work with respect to explainable AI planning and argumentation-based dialogues.

\subsection{Explainable AI Planning}

Explainable AI planning (XAIP) aims to foster trust, facilitate human-AI collaboration, and enable effective decision support in complex domains by providing users with understandable explanations of planning processes and decision-making \citep{Fox2017ExplainableP}. XAIP has been applied in various domains, including robotics \citep{setchi2020explainable}, healthcare \citep{saraswat2022explainable}, and beyond, highlighting its broad applicability and potential impact.

A central focus of XAIP research has been on the concept of model reconciliation \citep{chakraborti2017plan,sreedharan2021foundations,vas21,vasileioulogic,vasileioua2023please}, which seeks to align the mental models of AI agents and human users. However, the original framework often assumes that the agent has perfect knowledge of the human's model a priori, which can lead to incorrect assumptions and suboptimal explanations. To address this limitation, recent works have focused on relaxing the assumptions made about the human model. Notably, \citet{DungS22a} tackled this limitation from the perspective of answer set programming, tying their approach exclusively to planning problems. 

On the other hand, some related work tackled this limitation by considering uncertainty about the human’s model. In this context, \citet{sreedharan2018handling} propose a framework for reconciling with a set of possible human models, demonstrating how it can be used to provide explanations to multiple human users simultaneously. This work highlights the importance of accounting for the inherent uncertainty and variability in human mental models. In a related study, \citet{ijcai2018-671} developed a method for estimating the mental model from a provided foil, further emphasizing the need for techniques that can infer and update the agent's understanding of the human's model based on the available information.

\subsection{Argumentation-based Dialogues}

Argumentation-based dialogues have been developed to aid two (or more) agents in making decisions regarding their goals and plans. In this context, two agents with shared goals will only endorse plans that align with their beliefs. The literature on argumentation-based dialogues spans multiple disciplines, including AI \citep{bench2007argumentation}, legal reasoning \citep{zhong2014explaining}, and multi-agent systems \citep{nielsen2006generalization}, underlining the broad applicability and interdisciplinary nature of this research area.

\citet{belesiotis2010agreeing} propose an abstract argumentation-based protocol that enables two agents to deliberate on their proposals until they reach an agreement, guided by the persuasion-aligned planning beliefs of the agents. This work demonstrates the potential of argumentation-based approaches for facilitating collaborative decision-making and consensus-building in multi-agent settings. Argumentation-based explanation have also gained a lot of traction \cite{fan2015computing,shams2016normative,fan2018on,collins2019towards,oren2020argument,budan2020proximity,dennis2022explaining,Rago_23}. These works primarily focus on explanations whose justification is provided through argumentation semantics using specific dialogue formalizations, establishing an equivalence between the dialogues and the argumentation semantics. At the intersection of argumentation-based dialogues and XAIP is the work by \citet{vasileiou2023dr}, where the authors proposed a dialectical reconciliation dialogue between an AI agent and a human user with no assumptions about a known human model. The goal of this dialogue is to improve the understanding of the human user's understanding of the agent's decisions. While these approaches provide a solid foundation for argumentation-based explanations, they do not explicitly consider the uncertainty inherent in human-agent interactions. 

Most closely related to our setting, another line of research investigated probabilistic argumentation (\citep{hunter2013probabilistic,hunter2014probabilistic,hunter2022argument}) and introduces uncertainties in argumentation.
Specifically, uncertainties are represented by probabilistic measures, e.g., probabilities or degrees of belief, and assigned to propositions or arguments.
These works build upon probabilistic argument graphs, as defined in \citep{dung2010towards, li2011probabilistic}, providing a formal framework for reasoning about uncertain arguments.

% Closely related to our work is the area of probabilistic argumentation \citep{hunter2013probabilistic,hunter2014probabilistic,hunter2022argument}.
% Unlike the aforementioned studies, probabilistic argumentation considers the uncertainty in argumentation by representing it using probabilistic measures, such as probabilities or degrees of belief, assigned to propositions or arguments. These works build upon probabilistic argument graphs, as defined in \citep{dung2010towards, li2011probabilistic}, providing a formal framework for reasoning about uncertain arguments.

Building upon ideas from XAIP and argumentation-based dialogues, we provide a probabilistic approach to modeling and updating the agent's representation of the human's model.
Compared with previous works, our framework enables a more nuanced and adaptive approach to model reconciliation in XAIP by maintaining a probability distribution over possible mental models and updating it based on the human's trust and certainty feedback.

% To the best of our knowledge, we are the first to approximate a probabilistic human model during an argumentation-based dialogue.

% Our work builds upon ideas from XAIP and argumentation-based dialogues by introducing a probabilistic approach to modeling and updating the agent's representation of the human's model. 

% By maintaining a probability distribution over possible mental models and updating it based on human's trust and certainty feedback, our framework enables a more nuanced and adaptive approach to model reconciliation in XAIP. To the best of our knowledge, no other work has looked at approximating a probabilistic human model during an argumentation-based dialogue.

\section{Background}
We assume classical propositional logic for describing aspects of the world. We consider a finite (propositional) language $\fml{L}$ that utilizes the classical entailment relation, represented by $\models$. The set of \textit{models} (i.e. possible words) of $\fml{L}$ is denoted by $\fml{M}$, where each model $m_i \in \fml{M}$ is an assignment of true or false to the formulae of $\fml{L}$ defined in the usual way for classical logic. 
% {\color{blue}
% For $\phi \in \fml{L}$, $\texttt{Mod}(\phi)$ denotes the set of models of $\phi$, where $\texttt{Mod}(\phi) = \{m_i \in \fml{M} \: | \: m_i \models \phi \}$. In other words, the true assignment $m_i$ makes $\phi$ true. 
% }
For $\phi \in \fml{L}$, let~$\texttt{Mod}(\phi)=\{m_i \in \fml{M} \: | \: m_i \models \phi \}$ denote the set of models of $\phi$. 
% In other words, the true assignment $m_i$ makes $\phi$ true.

% {\color{blue}
% \begin{example}
%     Let $\fml{L}$ be the usual propositional formulae that can be formed from $\{a,b\}$, where $\fml{L} = \{a, a  \rightarrow b \}$. The set of models $\fml{M} = \{(a = \text{True}, b = \text{True}), (a = \text{True}, b = \text{False}), (a = \text{False}, b = \text{True}), (a = \text{False}, b = \text{False})\}$. In this way, for $\phi = a  \rightarrow b $, $\texttt{Mod}(\phi) = \{(a = \text{True}, b = \text{True}), (a = \text{False}, b = \text{True}), (a = \text{False}, b = \text{False})\}$.
%     \end{example}
% }

\subsection{Logic-based Argumentation}
We provide a partial review of logic-based argumentation~\citep{besnard2014constructing}. Our framework relies on an intuitive understanding of a logical \textit{argument}, which is essentially a set of formulae used to prove a specific claim.

\begin{definition}[Argument]\label{definition:argument}
Let $\fml{L}$ be the language and $\varphi \in \fml{L}$ a formula. An argument for $\varphi$ is defined as $A= \langle \Phi, \varphi \rangle$ such that: (i) $\Phi \subseteq \fml{L}$; (ii) $\Phi \models \varphi$; (iii) $\Phi \not \models \perp$; and (iv) $\nexists \Phi^{\prime} \subset \Phi$ s.t. $\Phi^{\prime} \models \varphi$.
\end{definition}

We refer to $\varphi$ as the \emph{claim} of the argument, and $\Phi$ as the \emph{premise} of the argument.

% We refer to $\varphi$ as the \emph{claim} of the argument, denoted as $\mathrm{C}(A)$, and $\Phi$ as the \emph{premise} of the argument, denoted as $\mathrm{P}(A)$. 
% The set of all arguments for a claim $\phi$ from $\KB$ is represented by $\mathcal{A}(\KB, \phi)$.

\begin{example}
Let $\fml{L}$ be a propositional language made up of variables $\{a, b, c, d, e \}$. Then,~$A_1=\langle\{a, b, a \wedge b \rightarrow c\}, c\rangle$ and $A_2=\langle\{b, d, d \rightarrow a, a \wedge$ $b \rightarrow c\}, c\rangle$ are two arguments for~$c$.
\end{example}

We incorporate a general definition of a \textit{counterargument} to address conflicting knowledge among agents. A counterargument is defined as an argument that opposes another argument by highlighting conflicts regarding the premises or claims. Specifically, 

\begin{definition}[Counterargument]
Let $\fml{L}$ be the language, and let $A_1=\langle\Phi, \varphi\rangle$ and $A_2=\langle\Psi, \psi\rangle$ be two arguments for $\varphi$ and $\psi$, respectively. We say that $A_2$ is a counterargument for $A_1$ iff $\Phi \cup \Psi \models \perp$.
\end{definition}

\begin{example}
Let $\fml{L}$ be a propositional language made up of variables $\{a, b, c, d, e \}$, and let $A_1 = \langle\{a, b, a \wedge$ $b \rightarrow c\}, c\}$ be an argument for $c$. Then, $A_{2}=$ $\langle\{f, d, f \wedge d \rightarrow \neg b\}, \neg b\rangle$ and $A_{3}=\langle\{e, e \rightarrow \neg c\}, \neg c\rangle$ are two counterarguments for $A_1$.
\end{example}

\subsection{Modeling Uncertainty in Propositional Logic}
% We then assume that we model uncertainty about the formulae using a probability distribution over models of the propositional logic.

Building on a propositional language $\fml{L}$, we can model the uncertainty of arbitrary formulae using a \textit{probability distribution} over the models $\fml{M}$ of $\fml{L}$. Formally,

\begin{definition}[Probability Distribution]
Let $\fml{M}$ be the set of models of the language $\fml{L}$. A probability distribution $P$ on $\fml{M}$ is a function $P: \fml{M} \mapsto [0,1]$ such that $\underset{m \in \fml{M}}{{\mathlarger{\sum}}}P(m) = 1$.
\end{definition}

In essence, a probability distribution over the models of $\fml{L}$ creates a \textit{ranking} between those models with respect to how likely they are to be true. This then allows us to quantify the uncertainty in a formula as follows:

\begin{definition}[Degree of Belief]
Let $\fml{M}$ be the set of models of language $\fml{L}$ and $P$ a probability distribution over $\fml{M}$. The degree of belief of a formula $\phi \in \fml{L}$ is $P(\phi) = \underset{m \models \phi}{{\mathlarger{\sum}}}P(m)$.
\end{definition}

We may refer to $P(\phi)$ as degree of belief or probability of $\phi$ interchangeably. Note that this approach to probabilities is essentially equivalent to probabilities assigned directly to the formulae~\cite{bacchus1989representing}.

\begin{example}
    Let $\fml{L}$ be a propositional language with variables $\{a,b\}$. An example of a probability distribution over the models $\fml{M}$ of $\fml{L}$ is shown in Table \ref{table:example_of_truth_table}. 
\begin{table}[h]
\centering
\begin{tabular}{|c|c|c|c|c|}
\hline
        & $m_1$ & $m_2$  & $m_3$  & $m_4$  \\ \hline
$a$       & True & True  & False & False \\ \hline
$b $      & True & False & True  & False \\ \hline
$P(m_i)$ & $0.1$ & $0.2$  & $0.4$  & $0.3$  \\ \hline
\end{tabular}
\caption{An example of probability distribution over models.}
\label{table:example_of_truth_table}
\end{table}
\end{example}
Then, $a$ has degree of belief $P(a) = P(m_1) + P(m_2) = 0.3$. Similarly, $a \rightarrow b$ has degree of belief $P(a  \rightarrow b) = P(m_1) + P(m_3) + P(m_4)  = 0.8.$.

\section{Approximating Human Models During Argumentation-based Dialogues}

% Effective argumentation requires understanding and reasoning about the knowledge of the dialogue participants. In human-AI interaction, the (AI) agent must typically maintain an estimate of the human's model in order to generate convincing arguments and meaningfully engage in dialogue. However, the agent is often uncertain about the human's knowledge state at the start of the interaction. 

In this section, we introduce a framework that allows an agent to progressively update its approximation of the human model over the course of an argumentation-based dialogue~\cite{vasileiou2023dr}. 

% Our approach leverages information revealed in the human's argumentative moves and utterances to refine a probabilistic estimate of their model.

\subsection{Problem Setting and Assumptions}
We consider an argumentation-based dialogue between two participants: an agent ($a$) and a human ($h$). We make the following key assumptions:

\squishlist
\item \textbf{Shared Domain Language}: Both $a$ and $h$ have access to and communicate in the same propositional language $\fml{L}$, with a shared vocabulary of atomic variables. This allows them to construct domain-specific formulae.

% \item \textbf{Agent Knowledge}: The agent's domain knowledge is encoded in a knowledge base $\KBa$ (i.e. a set of formulae). This knowledge base is not directly accessible to the human.

\item \textbf{Uncertain Human Model}: The agent maintains a probabilistic model of the human's knowledge, represented by a probability distribution $P_h$ over the possible models $\fml{M}$ of the human's knowledge at each step of the dialogue. This distribution captures the agent's uncertainty about the human's knowledge. Initially, the agent assumes a uniform prior $P_h^{t_0}(m) = \frac{1}{|\fml{M}|}$ for all $m \in \fml{M}$, representing agnosticism about the human model.

\item \textbf{Argument Traces}: We assume access to a (finite) argument trace $\fml{D} = \langle (A_1, x_1)^{t_1}, (A_2, x_2)^{t_2}, \ldots \rangle$ produced by the dialogue, where each $(A_i, x_i)^{t_i}$ represents an argument $A_i$ put forward by participant $x_i \in \{a, h\}$ at timestep $t_i$.

\squishend

% At any given time, an agent is assumed to consider a number of worlds possible. Among the possible worlds, some are more likely to be true than others. The probability distribution then allows the agent to have a description of the likelihood of each world. In addition, the distribution can be used to quantify the uncertainty in a formula

\subsection{Handling Argument Uncertainty}
In real-world argumentation, the arguments put forward by both the agent and the human often come with some degree of uncertainty. This uncertainty can arise from various sources, such as incomplete or imprecise knowledge, subjective interpretations, or lack of confidence in the reasoning process. In our framework, we consider two types of uncertainty associated with arguments:

\squishlist
\item \textbf{Uncertainty in Agent's Arguments}: When the agent presents an argument $A_i$ at timestep $t_i$, the human may not fully accept the argument as true, but rather evaluate it based on their trust in the agent. Intuitively, if the human has a high level of trust in the agent, they are more likely to assign a high probability to the agent's argument, indicating that they believe it is likely to be true or valid. Conversely, if the human has low trust in the agent, they may assign a lower objective probability to the argument, reflecting their doubts or skepticism about its correctness. We capture this notion of trust-based uncertainty using a trust value $\tau(A_i) \in [0,1]$ associated with each agent argument $A_i$, where higher values indicate greater trust.

\item \textbf{Uncertainty in Human's Arguments}: When the human puts forward an argument $A_j$ at timestep $t_j$, they may express some degree of uncertainty about it, represented by a probability $p(A_j) \in [0,1]$. This probability reflects the human's confidence in their own argument, based on factors such as their background knowledge, reasoning process, and awareness of potential counterarguments. Higher values of $p(A_j)$ indicate greater certainty in the argument.

\squishend

Now, according to prospect theory \cite{kahneman1979prospect}, the probability of an event may not align with the ``subjective'' perception of that probability, that is people tend to overweight small probabilities and underweight moderate to high probabilities when making decisions under uncertainty~\cite{fox2009prospect}.
To this end, we propose a probability weighting function \cite{gonzalez1999shape} to describe the relationship between the two.
In our scenario, we define the~\emph{trust value} of argument~$A_i$ as the human's subjective perception of the argument's uncertainty, denoted by~$\tau(A_i)$. To capture its relationship with the probability of the argument~$p(A_i)$, we use the following sigmoidal function~\cite{tversky1992advances}.

\begin{equation}
\tau(A_i) = \frac{p(A_i)^\gamma}{(p(A_i)^\gamma + (1-p(A_i))^{\gamma})^{1/\gamma}},
\label{eq:trust_to_prob}
\end{equation}
where $\gamma \in (0,1)$ is a parameter that controls the degree of this nonlinear distortion.

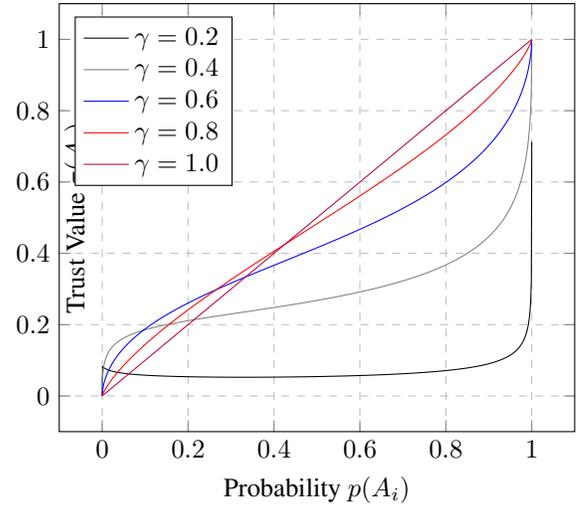
\begin{figure}[t]
\centering
\begin{tikzpicture}
\begin{axis}[
    legend pos=north west,
    y label style={at={(axis description cs:0.08,.5)},anchor=south},
    xlabel={Probability~$p(A_i)$},ylabel={Trust Value~$\tau(A_i)$},
    ymajorgrids=true,
    grid style=dashed,
    xmajorgrids=true,
    grid style=dashed,
    ]
\addplot [
    domain=0:1, 
    samples=1000, 
    color=black,
    ]
    {x^0.2/(x^0.2+(1-x)^0.2)^(1/0.2)};
    \addlegendentry{$\gamma=0.2$}
\addplot [
    domain=0:1, 
    samples=1000, 
    color=gray,
    ]
    {x^0.4/(x^0.4+(1-x)^0.4)^(1/0.4)};
    \addlegendentry{$\gamma=0.4$}
\addplot [
    domain=0:1, 
    samples=1000, 
    color=blue,
    ]
    {x^0.6/(x^0.6+(1-x)^0.6)^(1/0.6)};
    \addlegendentry{$\gamma=0.6$}
\addplot [
    domain=0:1, 
    samples=1000, 
    color=red,
    ]
    {x^0.8/(x^0.8+(1-x)^0.8)^(1/0.8)};
    \addlegendentry{$\gamma=0.8$}
\addplot [
    domain=0:1, 
    samples=1000, 
    color=purple,
    ]
    {x^1/(x^1+(1-x)^1)^(1/1)};
    \addlegendentry{$\gamma=1.0$}
\end{axis}
\end{tikzpicture}
\caption{Probability weighting function with $\gamma=0.2, 0.4,0.6, 0.8$ and~$1$.}
\label{fig:weighting_function} 
\end{figure}

\noindent Specifically, lower values of $\gamma$ (closer to 0) indicate excessive distortion (i.e., overweighting or underweighting) of the objective probability, while higher values (closer to 1) indicate a nearly linear relationship between the trust and the objective probability. The examples of relationships are shown in Figure \ref{fig:weighting_function}.

% \begin{figure}[h]
% \centering
% \includegraphics[width=0.4\textwidth]{figure/weighting_function.pdf}
% \caption{Relationship between Trust and Objective Probability with $\gamma = 0.5$.}
% \label{fig:weighting_function} 
% \end{figure}

The key idea behind using this function in our argumentation model is that it provides a psychologically plausible way to map the human's trust in an argument to an objective probability that is not just a linear scaling of trust. By accounting for the nonlinear biases in human probability assessment, the function allows the agent to more accurately model how the human is likely to respond to arguments of varying degrees of trustworthiness.

% Of course, the exact shape of the probability weighting function may vary across individuals and contexts. The $\gamma$ parameter can be tuned to fit empirical data on human probability perceptions in argumentative contexts. Moreover, alternative functional forms could be explored that capture other aspects of human probability bias. The proposed function is a parsimonious and interpretable starting point that is grounded in established psychological theory.

% By applying this probability weighting function, we can capture the psychological biases that humans exhibit when assessing probabilities under uncertainty. The function maps the human's trust $\tau(A_i)$ in an argument to a perceived probability $p(A_i)$ in a way that is consistent with empirical observations from prospect theory.

\subsection{Updating the Human Model}
We employ a Bayesian approach to update the agent's probability distribution $P_h$ over possible human models based on the uncertainties associated with the arguments exchanged during the dialogue. At each timestep $t_i$, when an argument $A_i$ is presented by either the agent or the human, we perform the following general update on the probability distribution:

\begin{equation}\label{eq:belief_update}
P_h^{t_i}(m) \!\! = \!\!\left\{
\begin{aligned}
& \frac{P_h^{t_{i-1}}(m)}{\sum_{m \models A_i} P_h^{t_{i-1}}(m)} \cdot p(A_i) & \text{if } m \models A_i  \\
& \frac{P_h^{t_{i-1}}(m)}{\sum_{m \not\models A_i} P_h^{t_{i-1}}(m)} \cdot (1-p(A_i)) & \text{if } m \not\models A_i
\end{aligned}
\right.
\end{equation}
where $m \models A_i$ denotes that model $m$ is consistent with argument $A_i$, i.e., the premises and conclusion of $A_i$ hold in $m$, and $p(A_i)$ is the probability associated with the argument. The proportionality constant is chosen to ensure that $P_h^{t_i}$ remains a valid probability distribution.

Intuitively, the update mechanism in Eq.~\eqref{eq:belief_update} increases the probability of human models that are consistent with the presented argument, weighted by the argument's associated probability $p(A_i)$. Models that are inconsistent with the argument have their probabilities decreased accordingly. The higher the probability of the argument, the more the distribution shifts towards consistent models.

Note that the probability $p(A_i)$ is determined based on the source of the argument. If $A_i$ is presented by the agent, $p(A_i)$ is the objective probability derived from the human's trust in the argument, $\tau(A_i)$, using the probability weighting function Eq.~\eqref{eq:trust_to_prob}. Specifically, $p(A_i)$ is obtained by numerically inverting Eq.~\eqref{eq:trust_to_prob} to solve for $p(A_i)$ given $\tau(A_i)$.

\begin{example}
Consider a dialogue where at timestep $t_1$, the agent asserts the argument $A_1 = \langle \{a, a \rightarrow b \}, \{b\} \rangle$. The human assigns a trust value of $\tau(A_1) = 0.6$ to this argument. Assuming $\gamma = 0.85$, the objective probability of $A_1$ is computed using Eq.~\eqref{eq:trust_to_prob}:

$$0.6 = \frac{p(A_1)^{0.85}}{[p(A_1)^{0.85}+(1-p(A_1))^{0.85}]^{\frac{1}{0.85}}}$$

Solving for $p(A_1)$, we get $p(A_1)\approx 0.62$. Suppose there are four possible models, $\mathcal{M} = \{m_1, m_2, m_3, m_4\}$, with a uniform prior distribution $P_h^{t_0}(m_1) = \ldots = P_h^{t_0}(m_4) = 0.25$. Let $m_1$ be the model that entails the premises of $A_1$, i.e., $m_1 \models \{ a, a \rightarrow b\}$. Applying the update mechanism from Eq.~\eqref{eq:belief_update}, we get:

\begin{equation*}
\begin{aligned}
 P_h^{t_1}(m_1) &= \frac{0.25}{0.25} \cdot 0.62 = 0.62 \\
 P_h^{t_1}(m_k) & =\frac{0.25}{0.25+0.25+0.25} \cdot 0.38 = 0.126 \; (k=2,3,4)
\end{aligned}
\end{equation*}

After this update, the model $m_1$ that is consistent with the agent's argument has a higher probability than the other three models, reflecting the human's moderate trust in the argument.
\end{example}

\noindent On the other hand, if $A_i$ is an argument presented by the human, $p(A_i)$ is the probability directly expressed by the human for their own argument.

\begin{example}
Continuing the previous example, suppose at timestep $t_2$, the human presents the argument $A_2 = \langle \{\neg a\}, \{\neg a \} \rangle$ with probability $p(A_2) = 0.9$.

Let $m_3$ and $m_4$ be the models that entail the premise of $A_2$. Applying the update mechanism to the distribution resulting from the previous (objective) probability update, we get:
\begin{equation*}
    \begin{aligned}
    P_h^{t_2}(m_1) &= \frac{0.62}{0.62+0.126} \cdot 0.1 = 0.083 \\
    P_h^{t_2}(m_2) &= \frac{0.126}{0.62+0.126} \cdot 0.1 = 0.017 \\
    P_h^{t_2}(m_3) &= P_h^{t_2}(m_4) = \frac{0.126}{0.126+0.126} \cdot 0.9 = 0.45 \\
    \end{aligned}
\end{equation*}

After this update, the models $m_3$ and $m_4$ that are consistent with the human's argument have much higher probability than the models consistent with the agent's previous argument.
\end{example}

By applying this update rule iteratively according to the sequence of arguments in the dialogue trace $\mathcal{D}$, the agent can gradually refine its estimate of the human model distribution $P_h$. The refined distribution incorporates information about the uncertainties associated with both the agent's and human's arguments, providing a more nuanced and psychologically grounded estimate of the human's knowledge.

It is worth noting that this update rule assumes that the human's trust in the agent's arguments and their own expressed probabilities are well-calibrated and consistent across the dialogue. In practice, there may be situations where the human's probability assessments are inconsistent or biased. Handling such inconsistencies and biases is an important challenge for future work. Nevertheless, the proposed update rule provides a simple and principled way to integrate argument uncertainties into the agent's modeling of the human's mental state, enabling more effective adaptation of the agent's argumentative strategies to the individual human.

\section{Empirical Evaluation: Human-User Study}

To evaluate the effectiveness of our proposed framework for approximating human models during argumentation-based dialogues, we conducted a user study simulating a real-world scenario. In this study, participants interacted with an AI assistant named ``Blitzcrank'' to assess the suitability of a fictional venue, ``Luminara Gardens'', for hosting a company team-building event. The study aimed to investigate the dynamics of human-AI interaction, the AI's ability to gauge participants' understanding, and changes in participants' trust levels throughout the dialogue.

Based on our proposed framework and the designed scenario, we formulated the following hypotheses:

\begin{quote}
    \textit{$H_1$: Our framework can effectively approximate a probability distribution that captures the participants' knowledge during argumentation-based dialogues.} \\

    \textit{$H_2$: Participants' trust in the AI assistant increases as the interaction progresses.}

\end{quote}

\subsection{Study Design}

\xhdr{Dialogue Design:} The study consisted of a series of interaction rounds between each participant and Blitzcrank. In each round, the participants were presented with a set of Blitzcrank's arguments regarding the suitability of Luminara Gardens for the team-building event. The arguments varied in their level of informativeness and persuasiveness, reflecting different degrees of argument strength.

After receiving an argument from Blitzcrank, the participants were asked to select their level of trust in the argument from four options: almost complete trust ($\tau = 0.9$), high trust ($\tau = 0.7$), average trust ($\tau = 0.5$), or low trust ($\tau = 0.2$). These trust levels $\tau(A_i)$ were mapped to objective argument probabilities $p(A_i)$ using the probability weighting function Eq.~\eqref{eq:trust_to_prob}. 
In our experiments, we computed~$p(A_i)$ numerically with the Newton–Raphson method~\cite{kelley2003solving}.
Table~\ref{tab:trust_levels} shows the computed probabilities with respect to the trust levels. Note that here we select $\gamma$ from the set $\{0.1, 0.2, 0.3, 0.4, 0.5, 0.6, 0.7, 0.8, 0.9\}$.

Next, the participants were presented with a set of five candidate counterarguments, each associated with a certainty level. The certainty level was inferred from linguistic cues in the arguments (e.g., ``I'm quite sure that...'', ``Is it possible that...''). The participant selected one of these counterarguments to present to Blitzcrank. Table~\ref{tab:certainty_levels} shows the classification of the participants' arguments by certainty level.

This process of argument presentation, trust assessment, and counterargument selection constituted one interaction round. After each round, the participants were asked to rank four different perspectives on Luminara Gardens' suitability for the event, based on their current understanding.\footnote{Note that the user model would be potentially changing over rounds, depending on the trust level of the agent's argument and the uncertainty level of the user's argument in the subsequent interactions.} These perspectives represented models, and the rankings provided a measure to assess the participant's understanding at each stage. Rank 1 indicated the most likely perspective (model), while rank 4 indicated the least likely one.

% \begin{table}[!t]
% \centering
% \resizebox{1.\columnwidth}{!}{ 
% \begin{tabular}{|l|c|c|}
% \hline
% \textbf{Trust Level} & \textbf{Trust Value} & \begin{tabular}[c]{@{}l@{}}\textbf{Objective Probability}\\ ( Eq.~\eqref{eq:trust_to_prob} with $\gamma = 0.85$)\end{tabular} \\ \hline
% Almost Complete Trust & $\tau = 0.9$ & 0.93\\ \hline
% High Trust & $\tau = 0.7$ & 0.74\\ \hline
% Average Trust & $\tau = 0.5$ & 0.51  \\ \hline
% Low Trust & $\tau = 0.2$ & 0.16  \\ \hline
% \end{tabular}
% }
% \caption{Mapping of trust levels selected by participants to objective probabilities of Blitzcrank's arguments.}
% \label{tab:trust_levels}
% \end{table}

\begin{table}[!t]
\centering
\resizebox{1.\columnwidth}{!}{ 
\begin{tabular}{|cc|c|c|c|c|}
\hline
\multicolumn{2}{|c|}{\textbf{Trust Level}}                                                                                        & \begin{tabular}[c]{@{}c@{}}Almost \\ Complete Trust\end{tabular} & \begin{tabular}[c]{@{}c@{}}High \\ Trust\end{tabular} & \begin{tabular}[c]{@{}c@{}}Average \\ Trust\end{tabular} & \begin{tabular}[c]{@{}c@{}}Low \\ Trust\end{tabular} \\ \hline
\multicolumn{2}{|c|}{\textbf{Trust Value}}                                                                                        & $\tau = 0.9$                                                     & $\tau = 0.7$                                          & $\tau = 0.5$                                             & $\tau = 0.2$                                         \\ \hline
\multicolumn{1}{|c|}{\multirow{6}{*}{\begin{tabular}[c]{@{}c@{}}\textbf{Objective} \\ \textbf{Probability}\end{tabular}}} & $\gamma = 0.4$ & 1.000                                                            & 0.990                                                 & 0.937                                                    & 0.150                                                \\ \cline{2-6} 
\multicolumn{1}{|c|}{}                                                                                  & $\gamma = 0.5$ & 1.000                                                            & 0.959                                                 & 0.804                                                    & 0.104                                                \\ \cline{2-6} 
\multicolumn{1}{|c|}{}                                                                                  & $\gamma = 0.6$ & 0.989                                                            & 0.898                                                 & 0.657                                                    & 0.114                                                \\ \cline{2-6} 
\multicolumn{1}{|c|}{}                                                                                  & $\gamma = 0.7$ & 0.972                                                            & 0.826                                                 & 0.566                                                    & 0.133                                                \\ \cline{2-6} 
\multicolumn{1}{|c|}{}                                                                                  & $\gamma = 0.8$ & 0.949                                                            & 0.765                                                 & 0.522                                                    & 0.155                                                \\ \cline{2-6} 
\multicolumn{1}{|c|}{}                                                                                  & $\gamma = 0.9$ & 0.922                                                            & 0.724                                                 & 0.504                                                    & 0.178                                                \\ \hline
\end{tabular}
}
\caption{Mapping of trust levels to the probabilities of Blitzcrank's arguments.}
\label{tab:trust_levels}
\end{table}

\begin{table}[!t]
\centering
\resizebox{1.\columnwidth}{!}{ 
\begin{tabular}{|l|c|l|}
\hline
\textbf{Certainty Level} & \textbf{Probability} & \textbf{Linguistic Cues} \\ \hline
High Certainty & 0.9 & \begin{tabular}[c]{@{}l@{}}``I am confident that...''\\ ``I am certain that...''\\ ``There is no doubt that...``\end{tabular} \\ \hline
Moderate Certainty & 0.7 & \begin{tabular}[c]{@{}l@{}}``It seems probable that...''\\ ``It’s quite likely that...''\\ ``There’s a good chance that...`` \end{tabular} \\ \hline
Neutral Uncertainty & 0.5 & \begin{tabular}[c]{@{}l@{}}``I’m not entirely sure, but...''\\ ``It could be the case that...''\\ ``There’s a possibility that...``\end{tabular} \\ \hline
Moderate Uncertainty & 0.3 & \begin{tabular}[c]{@{}l@{}}``There’s some doubt as to whether...''\\ ``It’s questionable whether...''\\ ``It’s uncertain if...``\end{tabular} \\ \hline
High Uncertainty & 0.1 & \begin{tabular}[c]{@{}l@{}}``I’m not confident in saying...''\\ ``It’s hard to say for sure...''\\ ``There’s significant uncertainty...''\end{tabular} \\ \hline
\end{tabular}
}
\caption{Classification of participant arguments by certainty level according to linguistic cues in their arguments.}
\label{tab:certainty_levels}
\end{table}

The dialogue continued for a fixed number of rounds (up to three) or until the participants chose to end the interaction. The specific arguments, counterarguments, and perspectives used in the study were designed to cover a range of aspects related to Luminara Gardens' suitability, such as venue capacity, catering options, entertainment facilities, and pricing.

\xhdr{Participant Details:} We recruited 150 participants via the Prolific platform \cite{palan2018prolific}. Participants were required to be fluent in English and were compensated $\$2.00$ for their time. Out of the 150 participants, 143 completed the study satisfactorily by passing attention checks and providing coherent responses.

Participants were divided into two groups based on the number of interaction rounds they chose to complete: Group A (44 participants) engaged in two rounds of interaction with Blitzcrank, while Group B (99 participants) engaged in three rounds, which was the maximum allowed. The choice of the number of rounds was up to the participants, reflecting their willingness to engage in a shorter or longer dialogue with the AI assistant.

\xhdr{Post-Study Questionnaire:} After completing the dialogue, participants answered a post-study questionnaire containing five Likert-scale items (1 - strongly disagree to 5 - strongly agree). Three items assessed changes in trust levels across the interaction rounds, while the remaining two items evaluated overall satisfaction with the interaction and the quality of Blitzcrank's arguments.

\subsection{Variants and Baselines}
In the following, we introduce three variants of our proposed method:
\squishlist
    % \item \emph{Variant 1: Personalization Upper Bound}: To achieve the overall best distribution of Spearman's rank correlation, the parameter $\gamma$ chosen in Eq.~\eqref{eq:trust_to_prob} varies for each participant.
    % Such a setting is reasonable due to distinct tradeoffs between trust level and objective probability among participants. 
    \item \emph{Variant 1: Personalization Upper Bound}: 
    Observe that every participant has a distinct relationship between his/her trust level and the objective probability.
    As such, we personalize the specific value of $\gamma$ in Eq.~\eqref{eq:trust_to_prob}, for every individual, that optimizes the distribution of Spearman's rank correlation using all user data in this variation.
    \item \emph{Variant 2: Personalization \uppercase\expandafter{\romannumeral1}}: (Group A) For each participant in Group A, we use the data from the first interaction process to determine the personal predicted value of $\gamma$, which is then applied to the second interaction process; (Group B) For each participant in Group B, we use the data from the first two interaction processes to determine the personal predicted value of  $\gamma$, which is then applied to the final interaction process.
    \item \emph{Variant 3: Personalization \uppercase\expandafter{\romannumeral2}}: Unlike Variant 2, for each participant in Group B, we use only the data from the first interaction process to determine the personal predicted value of $\gamma$, which is then applied to the last two interaction processes.
\squishend
To evaluate our framework, we compare our proposed method with the following three baselines:
\squishlist
    \item \emph{Baseline 1}: 
    The trust update does not apply the weighting function (i.e., $\gamma = 1$ in Eq.~\eqref{eq:trust_to_prob}). The probability distribution update involves assigning a probability to each model that entails the argument, followed by normalization. Formally,
    \begin{equation}\label{eq:belief_update_baseline1}
    P_h^{t_i}(m) \!\! = \!\!\left\{
    \begin{aligned}
    & \frac{p(A_i)}{Z} & \text{if } m \models A_i  \\
    & \frac{P_h^{t_{i-1}}(m)}{Z} & \text{if } m \not\models A_i
    \end{aligned}
    \right.
    \end{equation}
    where $Z = \underset{m \models A_i}{{\mathlarger{\sum}}}p(A_i) + \underset{m \not\models A_i}{{\mathlarger{\sum}}}P_h^{t_{i-1}}(m)$.
    \item \emph{Baseline 2}: The trust update does not apply the weighting function (i.e., $\gamma = 1$ in Eq.~\eqref{eq:trust_to_prob}). The probability distribution update follows Eq.~\eqref{eq:belief_update}.
    \item \emph{Baseline 3}: The trust update applies weighting function Eq.~\eqref{eq:trust_to_prob}. The probability distribution update follows Eq.~\eqref{eq:belief_update_baseline1}.
\squishend
Table \ref{table:baselines} shows the trust weighting rule and probability update of baseline methods.
\begin{table}[t]
\centering
\begin{tabular}{ccc}
\toprule
\textbf{Methods} & \textbf{Trust Weighting} & \textbf{Probability Update} \\ 
\midrule
Baseline 1          & $\gamma = 1$ in Eq.~\eqref{eq:trust_to_prob}   & Eq.~\eqref{eq:belief_update_baseline1}   \\
Baseline 2         & $\gamma = 1$ in Eq.~\eqref{eq:trust_to_prob}   & Eq.~\eqref{eq:belief_update}   \\
Baseline 3          & Eq.~\eqref{eq:trust_to_prob}       & Eq.~\eqref{eq:belief_update_baseline1}   \\
Proposed Method & Eq.~\eqref{eq:trust_to_prob} & Eq.~\eqref{eq:belief_update} \\
\bottomrule
\end{tabular}
\caption{Baseline methods}
\label{table:baselines}
\end{table}

\subsection{Evaluation Metrics}

To quantitatively evaluate our framework's performance in approximating human models and assess the significance of trust changes, we employed the following metrics:

\squishlist
\item \textbf{Spearman's Rank Correlation}: We computed Spearman's rank correlation coefficient \cite{spearmancorr} ($\rho$) between the participant's perspective rankings and the rankings generated by our framework at each interaction round. A high positive correlation indicates that our framework effectively approximates the participant's understanding of the situation.

\item \textbf{Student's $t$-Test}: To determine whether participants' trust levels increased between interaction rounds, we conducted paired $t$-tests \cite{student1908probable} with $p$-value $0.05$ comparing trust scores across rounds. Separate tests were performed for Group A (comparing trust between rounds 1 and 2) and Group B (comparing trust between rounds 1 and 2, and between rounds 2 and 3).
\squishend

\subsection{Results and Discussion}

The results of our user study provide strong support for both hypotheses $H_1$ and $H_2$.

\begin{figure}[t]
\centering
\includegraphics[width=0.45\textwidth]{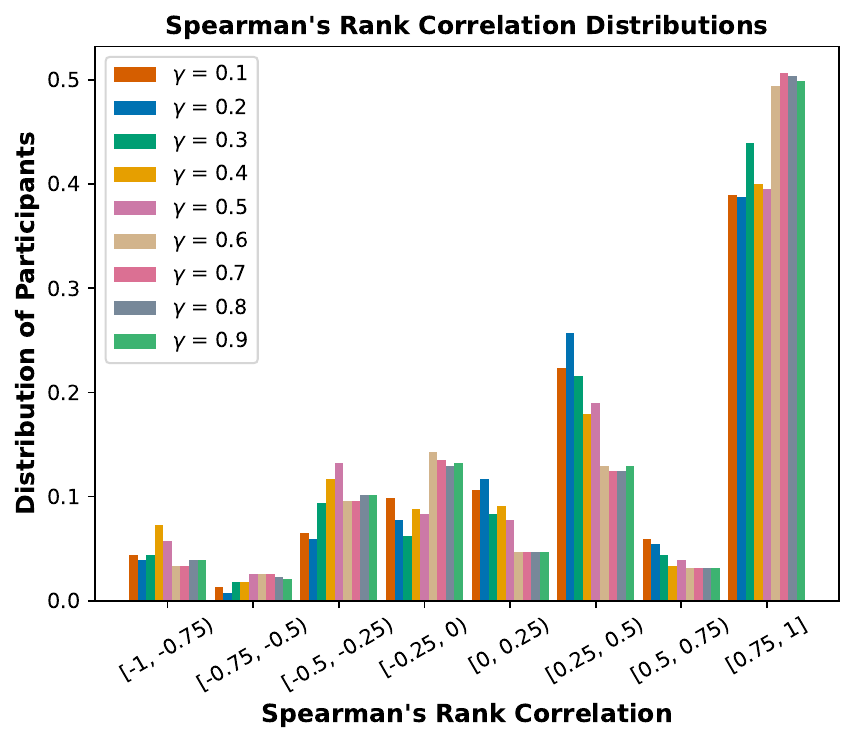}
\caption{Spearman's rank correlation distributions over different values of $\gamma$.}
\label{fig:parameter_spearman_rank_correlation} 
\end{figure}

Figure \ref{fig:parameter_spearman_rank_correlation} displays the distribution of Spearman's rank correlation coefficients over different values of $\gamma$ across all participants. The majority of coefficients are above $0.75$, indicating a substantial agreement between the participants' perspective rankings and those generated by our framework. This finding suggests that our approach effectively approximates a probability distribution that captures the participants' knowledge during argumentation-based dialogues, supporting hypothesis $H_1$. Moreover, $\gamma = 0.7$ produces the best distribution among all chosen values.

\begin{figure}[t]
\centering
\includegraphics[width=0.45\textwidth]{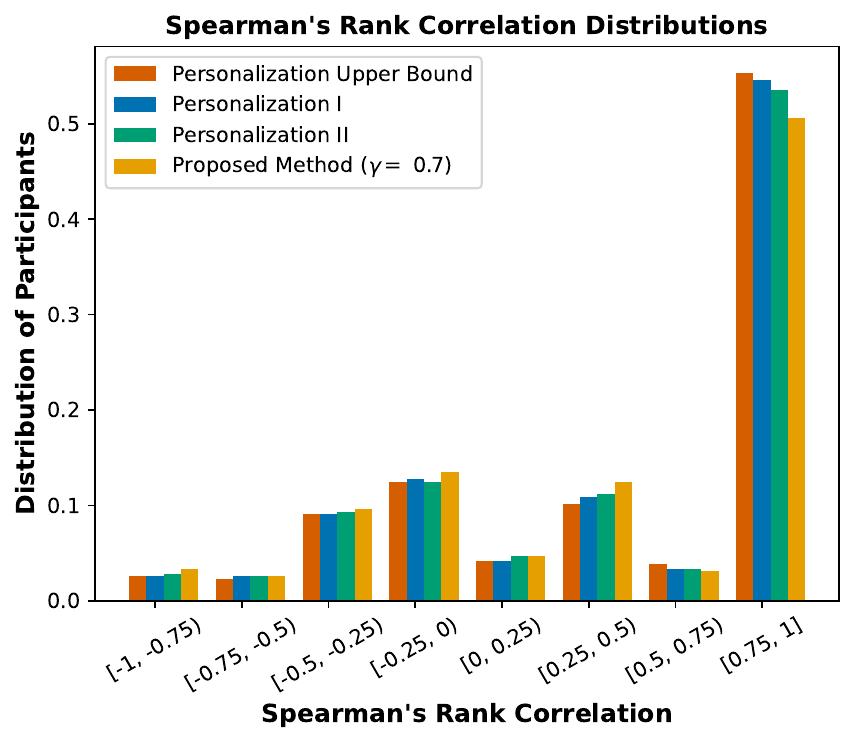}
\caption{Comparisons of Spearman's rank correlation distributions under different variants.}
\label{fig:baselines_spearman_rank_correlation} 
\end{figure}

Figure \ref{fig:baselines_spearman_rank_correlation} illustrates the distribution of Spearman's rank correlation coefficients for different variants. The Personalization Upper Bound achieves the optimal distribution of the proposed method. Moreover, Personalization \uppercase\expandafter{\romannumeral1} yields a better distribution than Personalization \uppercase\expandafter{\romannumeral2} because it utilizes more user data during the prediction process. Specifically, Personalization \uppercase\expandafter{\romannumeral1} can be considered to produce the best achievable distribution of our proposed method, given that obtaining the optimal distribution produced by the Personalization Upper Bound by using all user data for determining individual $\gamma$ values is unrealistic.
The effectiveness of all variants demonstrates the flexibility and potential of our proposed methods, thereby supporting hypothesis $H_1$.

\begin{figure}[h]
\centering
\includegraphics[width=0.45\textwidth]{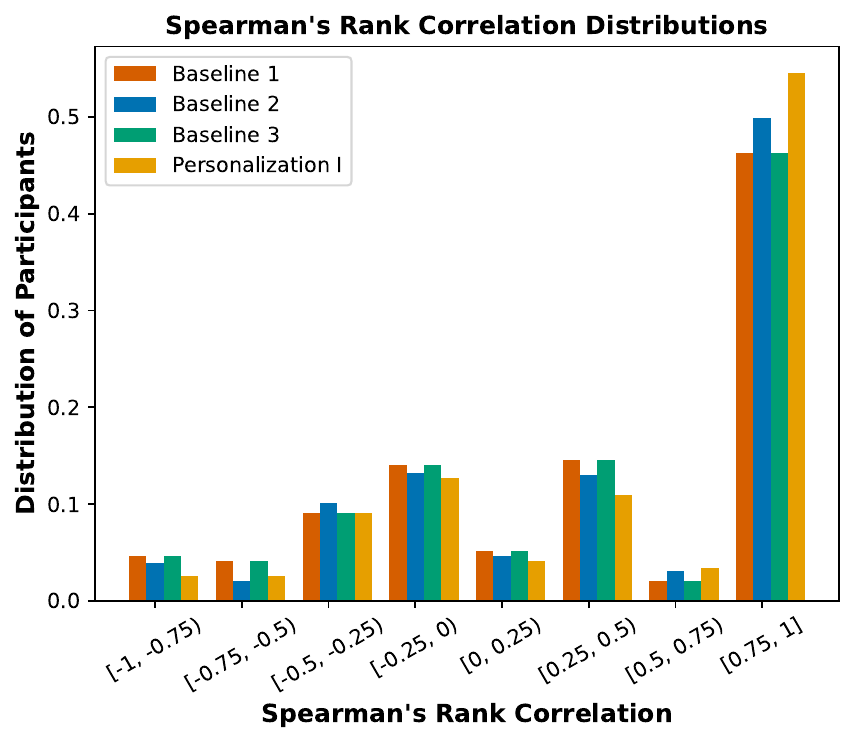}
\caption{Comparisons of Spearman's rank correlation distributions under different baselines. Note that Baselines 1 and 3 have the same distribution since the chosen values of $\gamma$ do not significantly influence the outcome due to the probability update function Eq.~\eqref{eq:belief_update_baseline1}.}
\label{fig:baselines_spearman_rank_correlation} 
\end{figure}

Figure \ref{fig:baselines_spearman_rank_correlation} shows that our proposed method outperforms the baseline methods in the distribution of Spearman's rank correlation coefficients, which hereby supports hypothesis $H_1$.
Specifically, our method enables~$55\%$ of correlation values to lie in the region of~$[0.75,1]$, exceeding the~$46\%,50\%,46\%$ provided by the baseline methods.

Table \ref{table:average_trust} shows a gradual increase in the average trust score as the dialogue proceeds. Besides, Table \ref{table:t-test} presents the results of the $t$-tests comparing trust scores between interaction rounds. For both Group A and Group B, we observe statistically significant increases in trust scores from round 1 to round 2 ($p_{1,2} < 0.05$). Additionally, for Group B, trust scores significantly increase from round 2 to round 3 ($p_{2,3} < 0.001$). Such results provide compelling evidence for hypothesis $H_2$, indicating that participants' trust in the AI assistant grows as the dialogue progresses and the assistant provides more relevant and persuasive arguments.

% Finally, Table~\ref{table:comprehension_and_satisfaction} shows the the post-study questionnaire responses further corroborate these findings, with participants reporting high levels of satisfaction with the interaction and the quality of Blitzcrank's arguments.

Our user study demonstrates the effectiveness of our proposed framework in an argumentation-based dialogue scenario. The results highlight the framework's ability to dynamically approximate a human model as a probability distribution, leading to increased trust and satisfaction among participants. Moreover, the post-study questionnaire responses further corroborate these findings, with participants reporting high levels of satisfaction with the interaction and the quality of Blitzcrank's explanations.

{\color{blue}
\begin{table}[!t]
\centering
\begin{tabular}{ccc}
\toprule
\textbf{Average Trust Score} & \textbf{Group A} & \textbf{Group B} \\ 
\midrule
First Round          & 0.511   & 0.508   \\
Second Round         & 0.634   & 0.552   \\
Third Round          & N/A        & 0.662   \\
\bottomrule
\end{tabular}
\caption{Average trust degree in different rounds.}
\label{table:average_trust}
\end{table}

\begin{table}[!t]
\centering
\resizebox{1.\columnwidth}{!}{ 
\begin{tabular}{lcc}
\toprule
& \textbf{Group A (2 rounds)} & \textbf{Group B (3 rounds)} \\
\midrule
$p_{1,2}$ & $0.00011297$ & $0.04303665$ \\
$p_{2,3}$ & N/A & $0.000001278$ \\
% \midrule
% \textbf{Stat. Significance} & \multicolumn{2}{c}{$p < 0.05$ indicates significance} \\
\bottomrule
\end{tabular}
}
\caption{Statistical significance ($p < 0.05$) of trust change. }
\label{table:t-test}

\end{table}
}

\section{Discussion and Conclusions}
In this paper, we introduced a novel framework for approximating human mental models during argumentation-based dialogues. Our approach leverages a Bayesian belief update mechanism to refine a probability distribution over possible human models based on the arguments exchanged throughout the dialogue. By incorporating uncertainty estimates for both the agent's and human's arguments, our framework provides a principled way to reason about the human's evolving knowledge state and perspectives.

The results of our human-subject study demonstrate the potential effectiveness of our framework in an applied argumentation setting. The high correlation between the rankings generated by our approach and the participants' actual perspective rankings suggests that our framework can capture some of the dynamics of human belief formation during argumentative interactions. 

% Moreover, the increases in participants' trust scores as the dialogue progressed indicate the framework's ability to adapt to individual users and provide increasingly persuasive arguments.

However, it is important to emphasize that this work is still in its early stages, and further research is needed to validate and refine the proposed framework. Our study had a limited sample size and focused on a single argumentation domain, so the generalizability of the findings to other contexts remains to be established. Moreover, the framework currently makes several simplifying assumptions, such as the consistency and calibration of human probability judgments, which may not hold in real-world settings.

One of the key aspects of our framework is the notion of argument-specific trust, which shapes the human's perception of the agent's arguments. We model trust as being influenced by factors such as perceived relevance, logical strength, consistency with prior beliefs, and clarity. The trust value assigned to each argument is treated as an observable input to the belief update process. However, our current framework does not explicitly model a global notion of trust, i.e., the human's overall trust in the agent across the entire dialogue. Extending the framework to incorporate a global trust component, and investigating its interplay with argument-specific trust, is an important direction for future work.

Another important consideration in practical argumentation systems is how to elicit the human's certainty levels for their own arguments. We envision several possible approaches, including explicit probability input, categorical confidence ratings, and inference from linguistic cues. A combination of these methods, allowing for both system-generated estimates and user adjustments, may provide a good balance of accuracy and usability. However, further empirical studies are needed to understand the impact of different elicitation methods on the quality and calibration of probability estimates in real-world settings.

In our study, the human users were asked to quantify their trust in the agent's argument~$A_i$ with a trust value~$\tau(A_i)\in[0,1]$. However, a human user may be uncertain about various parts of the argument. For example, let~$A_1=\langle\{a, a \rightarrow b\}, b\rangle$, and~$\tau(A_1) = 0.2$. This value does not indicate whether the user's uncertainty is on the conclusion of the argument (e.g., $b$) or in parts of its premises (e.g., $a$ or $a\rightarrow b$). Future work will look into a more nuanced argument uncertainty specification.

Another limitation of our current framework is the simplified representation of arguments as logical propositions. Real-world arguments often involve more complex structures and reasoning patterns, such as analogies, causal reasoning, and appeals to emotion. Capturing these rich argumentation dynamics within a computational framework is a significant challenge that requires further research at the intersection of argumentation theory, natural language processing, and knowledge representation.

In conclusion, the framework and study presented in this paper represent an initial step towards the development of adaptive, human-centric argumentation systems. While much work remains to be done to refine and validate our approach, we believe that this research direction has the potential to enhance the effectiveness of human-AI interaction.

\section*{Acknowledgments}

This research is partially supported by the National Science Foundation under award 2232055 and by J.P. Morgan AI Research. The views and conclusions contained in this document are those of the authors and should not be interpreted as representing the official policies, either expressed or implied, of the sponsoring organizations, agencies, or the United States government.

\bibliography{aaai24}

\end{document}